\documentclass[runningheads]{llncs}
\usepackage{graphicx}
\usepackage{amsmath,lipsum}
\usepackage{algorithm}
\usepackage[noend]{algpseudocode}
\usepackage{xspace}
\usepackage{adjustbox}
\usepackage{amsmath}

\usepackage{amsfonts}
\usepackage{xcolor}
\usepackage{booktabs}
\usepackage{multirow}
\usepackage{caption}
\usepackage{tablefootnote}

\newcommand{\track}{\textsc{Bio-ML}\xspace}

\usepackage{hyperref}
\usepackage{url}
\usepackage{breakurl}

\begin{document}
\title{
Machine Learning-Friendly 
Biomedical 
Datasets for
Equivalence and 
Subsumption Ontology Matching
}
%
%
%
\author{Yuan He\inst{1}
\and
 Jiaoyan Chen\inst{1}
 \and
 Hang Dong \inst{1}
 \and
 Ernesto Jim\'{e}nez-Ruiz  \inst{2,3}
 \and \\
 Ali Hadian \inst{4} \and 
 Ian Horrocks\inst{1}
 }

\titlerunning{ML-friendly Biomedical Datasets for Equivalence and Subsumption OM}
\authorrunning{He et al.}
%
\institute{Department of Computer Science, University of Oxford, UK
\email{\{yuan.he,jiaoyan.chen,hang.dong,ian.horrocks\}@cs.ox.ac.uk}
\and City, University of London, UK \\
\email{ernesto.jimenez-ruiz@city.ac.uk}
\and
University of Oslo, Norway
\and 
Samsung Research, UK\\
\email{a.hadian@samsung.com}
}
\maketitle              
\begin{abstract}

Ontology Matching (OM) plays an important role in many domains such as bioinformatics and the Semantic Web, and its research is becoming increasingly popular, especially with the application of machine learning (ML) techniques. 
Although the Ontology Alignment Evaluation Initiative (OAEI) 
represents an impressive effort for the systematic evaluation of OM systems, it still suffers from several limitations including limited evaluation of subsumption mappings, suboptimal reference mappings, and limited support for the evaluation of ML-based systems. To tackle these limitations, we introduce five new biomedical OM tasks involving ontologies extracted from Mondo and UMLS. Each task includes both equivalence and subsumption matching; the quality of reference mappings is ensured by human curation, ontology pruning, etc.; and a comprehensive evaluation framework is proposed to measure OM performance from various perspectives for both ML-based and non-ML-based OM systems. We report evaluation results for OM systems of different types to demonstrate the usage of these resources, all of which are publicly available as part of the new \track track at OAEI 2022.

\end{abstract}

{\small

\textbf{Resource type}: Ontology Matching Dataset

\textbf{License}: CC BY 4.0 International

\textbf{DOI}: \small\url{https://doi.org/10.5281/zenodo.6510086}

\textbf{Documentation}: \url{https://krr-oxford.github.io/DeepOnto/bio-ml/}


\textbf{OAEI track:} \url{https://www.cs.ox.ac.uk/isg/projects/ConCur/oaei/}

}

\section{Introduction}
Ontology Alignment (a.k.a. Ontology Matching (OM)) is the task of identifying inter-ontology entity pairs that are semantically related. A primary OM setting is matching \textit{named classes} with semantic \textit{equivalence} or \textit{subsumption} relationships, with the aim of integrating knowledge from different ontologies. 
A matched pair of named classes is known as an equivalence or subsumption mapping.
A successful OM case study is the Mondo Disease Ontology\footnote{\url{https://mondo.monarchinitiative.org/}} \cite{mondo}, which integrates disease concepts from various biomedical ontologies through mappings.
OM can also support interoperability among ontologies, and help to construct a unified terminology that extends the coverage of each individual ontology.
For example, given two classes about ``Desosom'' in the FMA (Foundational Model of Anatomy) ontology and the SNOMED CT (SNOMED Clinical Terms) ontology that are matched with equivalence, the subclass of the ``Desosom'' class named ``Autodesosome'' in FMA can be further inferred as a subclass of the ``Desosom'' class in SNOMED CT, thus augmenting SNOMED CT with more fine-grained knowledge. 
However, as ontologies evolve over time and become larger, it is unfeasible to have human beings annotating all the mappings;
hence (semi-)automatic OM systems are urgently needed \cite{onto-challenge}. 

Classical OM systems typically exploit text (e.g., labels and synonyms), structure (e.g., class hierarchies), and/or logical inference for class matching, and focus mostly on equivalence mappings.
For example, LogMap \cite{logmap} iteratively conducts
lexical matching, structure-based mapping extension and logic-based mapping repair; while AML \cite{aml} implements a matcher that considers various string-based heuristics, followed by mapping extension and repair.
Recently, machine learning (ML)-based OM systems have become increasingly popular as they can go beyond surface-form string comparison by encoding ontology entities into vectors. For example, DeepAlignment \cite{Kolyvakis2018DeepAlignmentUO} adopts \textit{counter-fitting} to refine word embeddings for better representation of class labels; VeeAlign \cite{Iyer2021VeeAlignMC} proposes a \textit{dual encoder} to encode both textual and path information of classes; and BERTMap \cite{He2021BERTMapAB} derives mappings through dynamic contextual text embeddings from the pre-trained language model BERT. 

For evaluation, the Ontology Alignment Evaluation Initiative\footnote{\url{http://oaei.ontologymatching.org/}} (OAEI) has been organizing a yearly evaluation campaign including several tracks (datasets) mainly comparing Precision, Recall, and F-score. 
Meanwhile, some recent OM studies, especially the ML-based ones, have proposed non-standard metrics and/or datasets with very incomplete gold standards. For example, Chen et al.\ used LogMap-ML~\cite{chen2021augmenting} to match the food ontology FoodOn with the health and lifestyle ontology HeLiS, and measured approximate Precision and Recall based on partial reference mappings, sampling and manual checking; and Neutel and de Boer~\cite{Neutel2021TowardsAO} measured coverage and MRR (Mean Reciprocal Rank) on industrial data with human judgement.

Despite the impressive community effort around the OAEI, the evaluation campaign still suffers from several limitations:
\begin{enumerate}
    \item \textbf{Limited evaluation metrics.} The prevalent evaluation metrics, Precision, Recall, and F-score, are of limited value when the reference mappings are incomplete, and can even stifle development by penalising advanced systems with high Recall that find correct mappings that are missing from the reference set.
    Some other metrics, such as approximate Precision and Recall based on sampling and human checking \cite{chen2021augmenting} or based on consensus by multiple systems \cite{harrow2017matching}, and Accuracy on distinguishing one positive mapping from few loosely constructed negative mappings \cite{nguyen2021biomedical}, may also be inaccurate and/or not sufficiently general.  
    
    \item \textbf{Suboptimal reference mappings.} 
    The reference mappings of many OM datasets are quite incomplete and/or incorrect. Such mappings are sometimes called \textit{silver standards} to distinguish them from (supposedly) complete gold standard mappings. The use of silver standards often leads to unfair comparisons among OM systems.
    For example, DeepAlignment \cite{Kolyvakis2018DeepAlignmentUO} exhibited better performance 
    than LogMap and AML when evaluated using silver standard mappings between the Schema.org and DBPedia ontologies. In this study AML achieved zero Recall, but closer examination of the results reveals that it actually retrieved some reasonable mappings that were not in the reference set. In the OAEI LargeBio track, some reference mappings are removed (or marked as ``ignore'') by an algorithm that repairs logical unsatisfiability resulting from the integration of the relevant ontologies~\cite{DBLP:conf/semweb/PesquitaFSC13}; 
    however, the mappings may still be correct according to human experts. 
    
    \item \textbf{Ignoring subsumption mappings.} The majority of existing resources are for equivalence matching. However, there are often more subsumption mappings than equivalence mappings between real-world ontologies, and the former could play an important role in knowledge integration and ontology curation. With the blooming research and application of ML and text understanding techniques, systems for subsumption matching (e.g., BERTSubs~\cite{chen2022contextual}) will likely become more feasible and widely investigated. 
    
    \item \textbf{Lack of support for ML-based Systems.}
    Most existing OM resources, including OAEI tracks, are not well suited to ML-based systems. They often do not consider hold-out validation sets required for tuning hyper-parameters (even non-ML-based systems may need such a validation set for adjusting parameters) and/or for 
    training in supervised or semi-supervised settings. 
    Moreover, during the development of an ML-based system,
    Precision, Recall and F-score are not very useful, 
    because computing the full output mappings is rather time-consuming and often does not directly reflect the capabilities of different ML modules or settings. 
    Ranking-based metrics are more suitable for ML development and are widely used in investigations of ML tasks such as knowledge graph completion \cite{Lin2015LearningEA,Rossi2021KnowledgeGE}, but they are rarely considered in the OM community. 
    These issues often lead to non-standardized and inconvenient evaluation set-ups for ML-based OM systems.

\end{enumerate}

To address the aforementioned issues,
in this paper, we present new large-scale OM resources based on Mondo and UMLS (Unified Medical Language System)\footnote{\url{https://www.nlm.nih.gov/research/umls/index.html}}, and propose a unified evaluation framework suitable for both ML-based and non-ML-based OM systems. 
This OM resource and evaluation framework is the basis for a new \track track in the OAEI 2022 campaign, which should be especially useful in attracting ML-based OM systems.
%
With Mondo, we create two OM tasks involving the OMIM (Online Mendelian Inheritance in Man), ORDO (Orphanet Rare Disease Ontology), NCIT (National Cancer Institute Thesaurus) and DOID (Human Disease Ontology) ontologies, which are tailored to the disease domain with high quality mappings curated by human experts. 
With UMLS, we use the semantic types (categories) of UMLS concepts to create multiple category-relevant tasks that involve the SNOMED CT, FMA, and NCIT ontologies. 
Briefly, our contributions can be summarized as follows:
\begin{enumerate}
    \item We have constructed an OM resource from Mondo which includes high quality manually curated mappings for the disease domain.
    \item We propose ontology \textit{pruning} to \textit{(i)} improve the relative completeness of reference mappings w.r.t.\ the pruned ontologies, and \textit{(ii)} obtain ontologies of various sizes to evaluate OM systems with different computational characteristics.
    In particular, for UMLS ontologies, we present a semantic-type-based pruning method for category-specific ontologies. 
    \item We have developed an approach to generate reference subsumption mappings from reference equivalence mappings. By deleting the classes involved in a given equivalence mapping,
    we ensure that the resulting subsumption mapping cannot be directly inferred from the equivalence mapping.
    \item We have formulated a unified evaluation framework which includes MRR (Mean Reciprocal Rank) and Hits@K as \textit{local ranking} metrics, which measure a system's ability to distinguish correct mappings from (non-trivial) false mappings; and Precision, Recall and F-score as \textit{global matching} metrics, which compare a system's final output mappings with the reference mappings.
    Our framework also includes standard data splitting: mappings are divided into validation and testing sets for unsupervised systems, and into training,  validation and testing sets for (semi-)supervised systems.

    \item We present preliminary evaluation results on our datasets for multiple OM systems of different types.
\end{enumerate}
    All the resources are open access, and we are setting up a new \track track within OAEI 2022 to promote their use and to attract more participation from the ML community.





\section{Resource Construction}

In this section, we introduce how our OM resources are constructed from the original ontology data shown in Table \ref{tab:ori_onto}. 
The resulting equivalence and subsumption matching datasets are presented in Table \ref{tab:pruned_onto} and \ref{tab:subs_onto}, respectively.

\begin{table}[!t]
    \centering
    \begin{adjustbox}{width=0.75\textwidth}
    \begin{tabular}{l l c c c c c }
    \toprule
    
      \textbf{Mapping Source} & \phantom{aa}  & \textbf{Ontology} & \phantom{aa} & \textbf{Ontology Source \& Version} & \phantom{aa} & \textbf{\#Classes} \\\midrule
         
        \multirow{4}{5em}{Mondo} 
         && OMIM & &Mondo\tablefootnote{Created from OMIM texts by Mondo's pipeline tool avaiable at:
       \url{https://github.com/monarch-initiative/omim}.} & & 44,729\\
         && ORDO & & BioPortal, \texttt{V3.2} & & 14,886\\
         && NCIT & & BioPortal, \texttt{V18.05d} & & 140,144 \\
         && DOID & & BioPortal, \texttt{2017-11-28} & & 12,498 \\
         
        \midrule
         
        \multirow{3}{5em}{UMLS} 
         && SNOMED & & UMLS, \texttt{US.2021.09.01}\tablefootnote{Created by the official snomed-owl-toolkit available at: \url{https://github.com/IHTSDO/snomed-owl-toolkit}, which keeps 350K classes of all the 490K classes in the original SNOMED CT.} & & 358,222 \\
         && FMA & & BioPortal, \texttt{V4.14.0} & & 104,523\\ 
         && NCIT & & BioPortal, \texttt{V21.02d} & & 163,842\\ 
          
    \bottomrule
    \end{tabular}
    \end{adjustbox}
    \vspace{-.15cm}
    \caption{\footnotesize Information of the source ontologies used for creating the OM resources.
    } 
    \label{tab:ori_onto}


    \begin{adjustbox}{width=0.96\textwidth}
    \begin{tabular}{l l c l c c c c c c c c }
    \toprule
    
         & \textbf{Ontology Pair} & \phantom{a} & \textbf{Category} & \phantom{a} &\textbf{\#Classes}& \phantom{a} & \textbf{\#Refs~($\equiv$)} & \phantom{a} & \textbf{\#Annot.} & \phantom{a} & \textbf{AvgDepths} \\\midrule
         
        \multirow{2}{5em}{Mondo} 
         & OMIM-ORDO && Disease & & 9,642-8,838 & & 3,721 & & 34K-34K & & 1.44-1.63 \\
         & NCIT-DOID && Disease &  &6,835-8,448 & & 4,684  & & 80K-38K & & 2.04-6.85\\
         
         \midrule
         
        \multirow{3}{5em}{UMLS} 
         & SNOMED-FMA && Body && 24,182-64,726 && 7,256 && 39K-711K && 1.86-9.32 \\ 
         & SNOMED-NCIT && Pharm && 16,045-15,250 && 5,803 && 19K-220K && 1.09-3.26 \\ 
         & SNOMED-NCIT && Neoplas && 11,271-13,956 && 3,804 && 23K-182K && 1.15-1.68\\
         
    \bottomrule
    \end{tabular}
    \end{adjustbox}
    \vspace{-.15cm}
    \caption{\footnotesize Statistics of each Mondo or UMLS equivalence matching task (dataset), including
    its two ontologies, its category (semantic type) for ontology pruning, its scale (named class and reference mapping sizes), the numbers of class annotations like labels, synonyms and definitions,
    and the average depth of named classes (depth is the minimum number of subclass hops from a named class to \texttt{owl:Thing})). ``Body'', ``Pharm'', and ``Neoplas'' denote semantic types of ``Body Part, Organ, or Organ Components'', ``Pharmacologic Substance'', and ``Neoplastic Process'' in UMLS, respectively.
    }
    \label{tab:pruned_onto}
    
    \begin{adjustbox}{width=0.68\textwidth}
    \begin{tabular}{l l c l c c c c}
    \toprule
    
         & \textbf{Ontology Pair} & \phantom{a} & \textbf{Category} & \phantom{a} &\textbf{\#Classes}& \phantom{a} & \textbf{\#Refs~($\sqsubseteq$)} \\\midrule
         
        \multirow{2}{5em}{Mondo} 
         & OMIM-ORDO && Disease & & 9,642-8,735 && 103  \\
         & NCIT-DOID && Disease &  &6,835-5,113 && 3,339  \\
         
         \midrule
         
        \multirow{3}{5em}{UMLS} 
         & SNOMED-FMA && Body && 24,182-59,567 && 5,506 \\ 
         & SNOMED-NCIT && Pharm && 16,045-12,462 && 4,225 \\ 
         & SNOMED-NCIT && Neoplas && 11,271-13,790 && 213 \\
         
    \bottomrule
    \end{tabular}
    \end{adjustbox}
    \vspace{-.15cm}
    \caption{\footnotesize Statistics of each Mondo or UMLS subsumption matching task (dataset),
    including 
    its two ontologies, its category (semantic type) for ontology pruning, its scales (named class and mapping sizes).
    Note that \textbf{\#Classes} of 
    the target ontology (right side) is smaller than the corresponding one in Table \ref{tab:pruned_onto} as some classes are deleted when constructing subsumption mappings.
    }
    \label{tab:subs_onto}
    \vspace{-0.8cm}
    
\end{table}

\subsection{Mondo Datasets} \label{mondo-dataset}
Our first two datasets are based on the cross-references in Mondo which is an integrated disease ontology with each of its classes matched to classes of some source ontologies \cite{mondo}. 
When constructing Mondo, curators first gathered reference mappings from various sources such as UMLS, MeSH (Medical Subject Headings), ICD (International Classification of Diseases). These mappings are deemed as semantically loose because there is no guarantee that they can be merged into a logically coherent ontology. Curators then adopted an ontology construction tool named k-BOOM to merge 
various source ontologies based on logical reasoning and Bayesian inference \cite{kboom}, and further invited domain experts for manual correction.
The merged ontology forms a more comprehensive terminology for rare diseases \cite{haendel2020}.

As suggested by the Mondo team, we selected two ontology pairs, OMIM-ORDO and NCIT-DOID, which are relatively up-to-date in Mondo.
OMIM is the primary online source of genes, genetic phenotypes, and gene-phenotype relations, based on manual curation from biomedical literature \cite{amberger2015omim}. 
The maximum class depth of the OMIM ontology is $2$, making it a typical example of ``flat'' ontology. Such ontologies have limited structural information, thus posing challenges to OM systems.
ORDO, the Orphanet Rare Disease Ontology,
includes a classification of rare diseases and relationships between diseases, genes and epidemiologic features; the ontology is derived from the Orphanet database, which is populated by literature curation and validated by international experts \cite{vasant2014ordo}. Many rare diseases are genetic disorders, therefore ORDO has a prominent overlap with OMIM, which is cross-referenced in ORDO and integrated in Mondo. NCIT (or NCIt) is a large ontology composed of various cancer-related concepts including cancer diseases, findings, drugs, anatomy, abnormalities, etc. \cite{nicholas2007ncit}, therefore it has a relatively smaller overlap with Mondo. DOID (or DO) stands for Human Disease Ontology, a regularly maintained source of human diseases \cite{Schriml2018do}, and most of its concepts are incorporated in Mondo. Matching NCIT and DOID 
will, in principle, identify the shared cancer-related diseases.
The versions of the selected ORDO, NCIT, and DOID ontologies (see Table \ref{tab:ori_onto}) are the closest to the most recent update of the Mondo mappings, according to Mondo's documentation\footnote{Mondo was working on official versioning, the information of current mappings is based on the preliminary release at: \url{https://github.com/monarch-initiative/mondo/tree/master/src/ontology/mappings}.}.
With the Mondo mapping data and these original ontologies, we create our OM datasets as follows:

\noindent\textbf{Ontology Preprocessing.} For each ontology, we conduct two preprocessing operations: \textit{(i)} removing obsolete or deprecated classes 
because they usually have up-to-date alternatives 
or are not in use anymore; \textit{(ii)} removing annotation properties that indicate cross-references to other data sources (e.g., \texttt{obo:hasDbXref}) because they could leak hints about the reference alignment to the OM systems. Unlike the  OAEI LargeBio track where some annotation properties are selected and merged into \texttt{rdfs:label}, we keep the rest of annotation properties and leave their interpretation to the OM systems.

\noindent\textbf{Ontology Pruning.}
Since the Mondo cross-references mainly aim at disease concepts, we prune each ontology by preserving disease classes and their contexts.
Specifically, if a class $c$ in an ontology is matched to a Mondo concept through the \texttt{skos:exactMatch} property, we preserve $c$; otherwise, we remove $c$ as well as all the axioms involving $c$, and at the same time directly assert its children as subclasses of each of its parents for keeping the hierarchy.
Pruning not only leads to OM tasks with ontologies of reasonable scale, but also improves the completeness of the reference mappings
w.r.t.\ the pruned ontologies.

\noindent\textbf{Equivalence Mapping Extraction.}
We extract equivalence mappings from the cross-references of each Mondo class, i.e., each pair of classes that are linked to the same Mondo class through the \texttt{skos:exactMatch} property is transformed to an equivalence mapping\footnote{We exclude mappings involving missing class ids.}. For example, \texttt{NCIT:C27518}\footnote{Compact IRI of a class in the form of \texttt{ontology\_prefix:class\_ID}.} and \texttt{DOID:4321} form an equivalence mapping because they are both mapped to the Mondo concept, \texttt{MONDO:0002961} (``Large Cell Acanthoma''). 

\noindent\textbf{Subsumption Mapping Extraction.}
We construct subsumption reference mappings based on the  equivalence reference mappings. 
Given an equivalence mapping ($c$, $c'$), we extract a subsumption mapping ($c$, $c''$) where $c''$ is an asserted subsumer of $c'$ in the ontology of $c'$.
Taking the example of \texttt{DOID:4321} (``Large Cell Acanthoma''), which is equivalently matched to \texttt{NCIT:C27518}; since one of its parent classes is \texttt{DOID:174} (``Acanthoma''), a potential subsumption mapping is (\texttt{NCIT:C27518}, \texttt{DOID:174}).
Note that both $c$ and $c'$ could have multiple asserted and inferred subsumers; considering all of them could lead to excessively many subsumption mappings for each equivalence mapping. 
Our solution simply selects one of the most specific subsumers of $c'$. This leads to challenging but incomplete subsumption mappings.
Thus, when evaluating subsumption matching, we do not consider Recall (see Section \ref{sec:evaluation_framework}).
To evaluate a system's ability on directly inferring cross-ontology subsumptions, we prevent it from utilizing the original equivalence mapping ($c$, $c'$) by deleting $c'$. 
As in ontology pruning, after deleting $c'$, its parent classes are asserted to be subsumers of each of its child classes, so as to preserve the class hierarchy.
It is possible that the deleted class appears in some other equivalence mappings to process, or some subsumption mappings that have been created. 
For the former, we skip such equivalence mappings, while for the later, we remove such subsumption mappings.

\subsection{UMLS Datasets}

UMLS is one of the most comprehensive mapping efforts, and integrates over $200$  vocabularies to create a biomedical metathesaurus \cite{Bodenreider2004}. 
As an integrated ontology, it incorporates well-known ontologies such as SNOMED CT, FMA, NCIT, and 
GO (Gene Ontology). 
It describes millions of biomedical concepts, and relationships among them. 
Each concept is classified into one or more hierarchical \textit{semantic types} (or categories) such as ``Finding'', ``Chemicals'', and  ``Substance''.

To construct OM datasets from UMLS, we selected its latest version \texttt{2021AB} at the time of doing experiments, and 
downloaded three of its corresponding ontologies --- SNOMED CT, FMA and NCIT, all of which are large biomedical ontologies with over 100K named classes (see Table \ref{tab:ori_onto} for more information).
SNOMED CT\footnote{The license to access UMLS is global and can be used to access SNOMED CT. We obtained SNOMED CT (and UMLS) after signing up to the UTS account and license following SNOMED and UMLS licensing in \url{https://www.nlm.nih.gov/healthit/snomedct/snomed_licensing.html}.} has a more general and comprehensive coverage of clinical terms to support electronic healthcare systems and clinical applications \cite{chap23HIbookCoiera,donnelly2006snomed}, while FMA \cite{rosse2008fma} and NCIT (as introduced previously) are mainly about human anatomy and cancer, respectively. 

We first performed the same preprocessing as described in Section \ref{mondo-dataset}, and then established category-specific alignment tasks by pruning the ontologies via semantic types, i.e., we preserve classes of a chosen semantic type, delete the other classes, and preserve the hierarchy of the superclasses and subclasses of each deleted class as in ontology pruning for Mondo.
The equivalence reference mappings are extracted from cross-ontology classes that are matched to the same UMLS concept \cite{jimenez2011logic}, and the subsumption reference mappings are constructed from the equivalence mappings in the same way as for Mondo. 

\section{Evaluation Framework}\label{sec:evaluation_framework}

We propose a comprehensive OM evaluation framework with different metrics of \textit{local ranking} and \textit{global matching} under both  \textit{unsupervised} (fully automatic) and \textit{semi-supervised} settings. 
Metrics of local ranking are to measure a system's capability on distinguishing true mappings and (hard) false mappings; while metrics of global matching are to measure whether a system can output a set of mappings close to the reference mappings.
The semi-supervised setting enables the evaluation of some ML-based systems that require training mappings.

\subsection{Local Ranking}

Given a reference mapping $m$=($c$, $c'$), where $c$ and $c'$ are two classes from the to-be-aligned ontologies $\mathcal{O}$ and $\mathcal{O}'$, respectively,
an OM system is required to distinguish $m$ from its corresponding set of negative mappings (denoted as $\mathcal{M}_m$) by assigning $m$ with a higher matching score. $\mathcal{M}_m := \left\{(c, c'') | c'' \in \mathcal{C}_{neg} \right\}$ is constructed by combining $c$ with a set of mismatched (negative) candidate classes (denoted as $\mathcal{C}_{neg}$)  from $\mathcal{O}'$.
With the mapping scores, we adopt ranking-based evaluation metrics $Hits@K$ ($H@K$ in short) and $MRR$ (Mean Reciprocal Rank), which are computed as follows:
$$
Hits@K = \frac{|m \in \mathcal{M}_{ref} \ | \ Rank(m) \leq K)\}|}{|\mathcal{M}_{ref}|}
$$
$$
MRR = \frac{\sum_{m \in \mathcal{M}_{ref}} Rank(m)^{-1}}{|\mathcal{M}_{ref} |}
$$
where $\mathcal{M}_{ref}$ denotes the set of reference mappings, 
$Rank(m)$ returns the ranking position of $m$ among $\mathcal{M}_m \cup \left\{m\right\}$ according to their scores,
$K$ (often set to $1$, $5$ and $10$) denotes the ranking position that is concerned.
We could consider all the classes in $\mathcal{O}'$ for constructing $\mathcal{M}_m$, but this frequently results in excessive evaluation time, especially for large-scale ontologies.
To ensure the evaluation efficiency, which is particularly important for ML-based model comparison/selection, we sample challenging negative candidates with heuristics introduced as follows.
\vspace{-.1cm}

\subsubsection{Negative Candidate Generation.}
Given a reference mapping $m$=($c$, $c'$), we consider three strategies to construct $\mathcal{C}_{neg}$ from $\mathcal{O}'$.
\begin{enumerate}
    \item \textbf{IDFSample} (text similarity-based). This strategy is to introduce hard negative candidates that are ambiguous to the ground truth class at text level (i.e., with similar labels\footnote{\label{note:labels}Labels are extracted from annotation properties concerning synonyms of the class name, e.g., \texttt{rdfs:label}, \texttt{fma:synonym}, \texttt{skos:prefLabel}, etc.}). We first build a sub-word inverted index \cite{He2021BERTMapAB} for the labels of all the classes of $\mathcal{O}'$ using a sub-word tokenizer pre-trained on biomedical texts \cite{Alsentzer2019PubliclyAC}. With this index, we select top-$N$ classes from $\mathcal{O}'$ according to the $idf$ (inverted document frequency) scores in descending order:
    $$s(c', c'') = \sum_{t \in Tok(c') \cap Tok(c'')} \log_{10} \frac{|C'|}{|I(t)|}$$
    where $Tok(\cdot)$ gives all sub-word tokens of a class's labels, $I(t)$ returns classes of $\mathcal{O}'$ whose labels contain the token $t$, and $C'$ denotes all the classes of $\mathcal{O}'$.
    
    \item \textbf{NeighbourSample} (graph context-based). This strategy is to introduce hard negative candidates that are close to the ground truth class along class hierarchy. With the asserted subsumption axioms in an ontology, we can establish an \textit{undirected} graph with named classes as nodes and subclass (\texttt{rdfs:subClassOf}) relations as edges. We adopt breadth-first search (BFS) over the subclass edges (bidirectional)  to add the neighbouring classes of $c'$ as candidates. 
    The search starts from one-hop away neighbours, then goes to two-hop away neighbours, and so forth. It terminates when the number of neighbours (candidates) exceeds the required number $N$ or the preset maximum number of hops has been reached. It is possible to obtain more than $N$ candidates by adding all $r$-hop away neighbours; in this case, we sample among these $r$-hop candidates randomly to meet the number. 
    Note that we exclude the root class \texttt{owl:Thing} from BFS.
    This restricts the candidates within the branch of $c'$, leading to high quality negative candidates and significantly improving the searching efficiency. 
    
    \item \textbf{RandomSample}. This strategy is to randomly select negative candidates from the classes of $\mathcal{O}'$, as a complement to the above two strategies.
    
\end{enumerate}

\begin{algorithm}[!t]
\footnotesize
\caption{\small Negative Candidate Generation}
\label{alg:neg_cand_gen}
\textbf{Input}: A reference mapping,  $m=(c, c')$; Generation strategies $\{\mathcal{S}_1, \mathcal{S}_2, ..., \mathcal{S}_n\}$, and their corresponding numbers of negative candidates to generate $\{N_1, N_2, ..., N_n\}$\\
\textbf{Output}: Negative candidates for $m$, $\mathcal{G}(m)$\\
\vspace{-0.4cm}
\begin{algorithmic}[1] 
\State $\mathcal{T}(m) \gets$ invalid candidates for $m$
\State Initialize the set of negative candidates: $\mathcal{G}(m) \gets \{ \}$
\For{$i \gets 1 \text{ to } n$}
\State Generate unique $|\mathcal{G}(m)| + |\mathcal{T}(m)| + N_i$ raw samples with strategy $\mathcal{S}_i$ as $\mathcal{G}_i(m)$
\State Remove those have been sampled and invalid: $\mathcal{G}_i(m) \gets \mathcal{G}_i(m) \ \backslash \ (\mathcal{G}(m) \cup \mathcal{T}(m))$
\State Truncate $\mathcal{G}_i(m)$ to first $N_i$ (ranked) samples if $| \mathcal{G}_i(m) |> N_i$ 
\While{$| \mathcal{G}_i(m) | < N_i$}
\State Randomly select $N_i - | \mathcal{G}_i(m) |$ unique candidates as $\mathcal{R}$
\State $\mathcal{G}_i(m) \gets (\mathcal{G}_i(m) \cup \mathcal{R}) \ \backslash \ (\mathcal{G}(m) \cup \mathcal{T}(m))$
\EndWhile
\State $\mathcal{G}(m) \gets \mathcal{G}(m) \cup \mathcal{G}_i(m)$
\EndFor
\State \textbf{return} $\mathcal{G}(m)$
\end{algorithmic}
\end{algorithm}
\normalsize
To ensure we always get the required number of negative candidates with no duplicates, 
we combine the above three strategies with a Negative Candidate Generation algorithm (see Algorithm \ref{alg:neg_cand_gen}) which has the following characteristics: 
\begin{enumerate}
    \item The above strategies could occasionally generate positive candidates, i.e., classes that can be matched to $c$. These classes are pre-computed (in Line~1, denoted as $\mathcal{T}(m)$) and excluded from negative candidates.
    For subsumption matching, $\mathcal{T}(m)$ further incorporates the asserted and inferred subumers of $c'$ since their combinations with $c$ are not negative subsumption mappings.
    \item At $i^{th}$ iteration, only when strategy $\mathcal{S}_i$ cannot generate $N_i$ ($N_i << |C'|)$  \textit{new} candidates will RandomSample be used to amend the number. The reason for sampling $|\mathcal{G}(m)| + |\mathcal{T}(m)| + N_i$ raw candidates first (in Line 3; the current set of negative candidates is denoted as $\mathcal{G}(m)$) is that in the worst case scenario, all the generated candidates are either duplicated or invalid. Therefore, the algorithm samples $|\mathcal{G}(m)| + |\mathcal{T}(m)|$ more than required first to preserve as many candidates as possible.
\end{enumerate}
Overall, for each reference mapping $m=(c, c')$, we sample $\sum_{i=1}^n N_i$ (defined in Input of Algorithm \ref{alg:neg_cand_gen}) unique negative candidates and add $c'$ as the only positive candidate; we then compute the ranking-based metrics for each OM system that supports class pair (mapping) scoring.

\subsection{Global Matching}

To eventually determine the output mappings, an OM system requires not only a mapping scoring module (which can be evaluated by local ranking), but also other components such as mapping searching, blocking, extension and repair.
The prevalent metrics for measuring the final output mappings
are Precision ($P$), Recall ($R$), and F-score:

$$
    P = \frac{|\mathcal{M}_{out} \cap \mathcal{M}_{ref}|}{|\mathcal{M}_{out}|}, \ \
    R = \frac{|\mathcal{M}_{out} \cap \mathcal{M}_{ref}|}{|\mathcal{M}_{ref}|}, \ \
    F_{\beta} = (1+\beta^2) \cdot \frac{P \cdot R}{\beta^2 \cdot P + R} 
$$
where $\mathcal{M}_{out}$ and $\mathcal{M}_{ref}$ correspond to mappings computed by an OM system and the reference mappings, respectively; $\beta$, often set to $1$, is a weighting for Precision and Recall. 
%
The global matching evaluation can demonstrate the overall performance of an OM system, but it is not well applicable for developing the ML-based mapping scoring module that has been widely considered in OM research in recent years, since \textit{(i)} the output mappings depend on several other modules besides mapping scoring, and \textit{(ii)} computing all the mappings is rather time-consuming (the naive traversal has a quadratic mapping search space), leading to very inefficient evaluation for ML models.
Meanwhile, when the reference mappings are incomplete, we are essentially penalizing OM systems with good Recall.
The local ranking evaluation can address these issues and thus, it is a good complement to the global matching evaluation. 


\subsection{Data Splitting}

For both evaluation schemes, we consider two settings for reference mapping splitting. 
The first setting splits the reference mappings into $10\%$ hold-out validation set 
for hyperparameter tuning or model selection, and $90\%$ testing set 
for final evaluation. 
Such setting can be used for comparing fully automatic non-ML-based OM systems and unsupervised ML-based OM systems. 
The second setting splits the reference mappings into  $20\%$, $10\%$, and $70\%$, corresponding to training, validation, and testing sets, respectively. 
Such a setting can evaluate those ML-based OM systems that are able to (or have to) use a small portion of given mappings for training. 
Note that the prevalent (fully) supervised learning data split with large portion of training data
is not applicable for OM because of the extreme positive-negative imbalance, i.e., the number of correct mappings is of several orders smaller than the incorrect ones.

It is worth mentioning when calculating Precision, Recall and F-score on a particular set of the reference mappings, we need to  
\textit{exclude} reference mappings that are not in this set from the system output mappings;
e.g., Precision on the testing set\footnote{Note that during validation we cannot compute Precision, Recall, and F-score directly because the complete mapping set is not available at this stage. However, we can still use validation mappings to tune hyperparameters using alternative metrics such as accuracy, Hits@K, MRR, etc.} $\mathcal{M}_{test}$ is computed as: 
$$
P_{test} = \frac{|\mathcal{M}_{out} \cap \mathcal{M}_{test}|}{|\mathcal{M}_{out} \ \backslash \ (\mathcal{M}_{ref} \ \backslash \ \mathcal{M}_{test}) |}
$$





\section{Evaluation Results}

\subsection{Equivalence Matching}

For equivalence matching, we evaluated the following OM systems (methods):
\begin{enumerate}
    \item \textbf{EditSim\footnote{\label{deepontourl}EditSim and BERTMap codes: \url{https://github.com/KRR-Oxford/DeepOnto}}}. Many of the equivalent concepts have a similar naming and therefore, measuring class similarity based on simple edit distance between class labels is a reasonable baseline. Specifically, this method computes the matching score between two classes using the maximum of the normalized edit similarity scores among the combinations of their labels\footnotemark[11].
    Note that the normalized edit similarity score is defined as \textit{1 $-$ normalized edit distance}.
    \item \textbf{LogMap\footnote{\url{https://github.com/ernestojimenezruiz/logmap-matcher}} \& AML\footnote{\url{https://github.com/AgreementMakerLight/AML-Project}}}. 
    LogMap and AML are two classical OM systems based on lexical matching, mapping extension and repair. They are leading OM systems in many equivalence matching tasks including those in the~OAEI.
    \item \textbf{BERTMap\footnotemark[15]}. BERTMap  is a ML-based OM system which uses class labels\footnotemark[11] to fine-tune a pre-trained language model for synonym classification, and then aggregates the synonym scores as the mapping score. 
    For efficient prediction,
    it exploits the sub-word inverted index for candidate selection and uses EditSim to filter mappings whose two classes have a common class label. 
    Note that we employ the same candidate selection method for EditSim. 
\end{enumerate}
The validation set is used for tuning hyperparameters such as the mapping filtering threshold 
of BERTMap and EditSim, and the selection of annotation properties. 
The numbers of negative candidates using IDFSample and NeighbourSample are both set to $50$, and RandomSample is used only for compensating the number. 
In total, for each reference mapping, the systems need to rank $100$ negative candidates plus $1$ ground truth class.

\begin{table*}[!t]
	\centering
    \begin{adjustbox}{width=0.85\textwidth}
	\begin{tabular}{l l c c c c c c c c c c c c c c c c c c}
		\toprule
		 & & & \multicolumn{6}{c}{\textbf{90\% Test Mappings}} & \phantom{a} & \multicolumn{6}{c}{\textbf{70\% Test Mappings}} \\
		 \cmidrule{4-6} \cmidrule{8-9} \cmidrule{11-13} \cmidrule{15-16} 
		 \textbf{Task} & \textbf{System} & &  \textbf{P} & \textbf{R} & \textbf{F1} && \textbf{MRR} & \textbf{H@1} && 
		 \textbf{P} & \textbf{R} & \textbf{F1} && \textbf{MRR} & \textbf{H@1} \\ \midrule
		 \multirow{4}{5em}{\footnotesize OMIM-ORDO (Disease)} & EditSim && 0.819 & 0.499 & 0.620 && 0.776 & 0.729 && 0.781 & 0.507 & 0.615 && 0.777 & 0.727 \\
		 & LogMap && 0.827 & 0.498 & 0.622 && 0.803 & 0.742 && 0.788 & 0.501 & 0.612 && 0.805 & 0.744 \\
		 & AML && 0.749 & 0.510 & 0.607 && NA & NA && 0.702 & 0.517 & 0.596 && NA & NA\\
		 & BERTMap && 0.730 & 0.572 & 0.641 && 0.873 & 0.817  && 0.762 & 0.548 & 0.637 && 0.877 & 0.823\\ \midrule
		 \multirow{4}{5em}{\footnotesize NCIT-DOID (Disease)} & EditSim && 0.912 & 0.776 & 0.838 && 0.904 & 0.884 && 0.889 & 0.771 & 0.826 && 0.903 & 0.883 \\
		 & LogMap && 0.918 & 0.667 & 0.773 && 0.559 & 0.364 && 0.896 & 0.661 & 0.761 && 0.559 & 0.363 \\
		 & AML && 0.873 & 0.773 & 0.820 && NA & NA && 0.841 & 0.770 & 0.804 && NA & NA\\
		 & BERTMap && 0.912 & 0.829 & 0.868 && 0.967 & 0.953 && 0.823 & 0.887 & 0.854 && 0.968 & 0.955 \\ \midrule
		 \multirow{4}{5em}{\footnotesize  SNOMED-FMA (Body)} & EditSim && 0.976 & 0.660 & 0.787 && 0.895 & 0.869 && 0.970 & 0.665 & 0.789 && 0.897 & 0.871 \\
		 & LogMap && 0.702 & 0.581 & 0.636 && 0.545 & 0.330 && 0.646 & 0.580 & 0.611 && 0.542 & 0.328 \\
		 & AML && 0.841 & 0.776 & 0.807 && NA & NA && 0.805 & 0.779 & 0.792 && NA & NA\\
		 & BERTMap && 0.997 & 0.639 & 0.773 && 0.954 & 0.930 && 0.811 & 0.708 & 0.756 && 0.967 & 0.950\\ \midrule
		 \multirow{4}{5em}{\footnotesize SNOMED-NCIT (Pharm)} & EditSim && 0.979 & 0.432 & 0.600 && 0.836 & 0.760 && 0.973 & 0.429 & 0.595 && 0.835 & 0.758\\
		 & LogMap && 0.915 & 0.612 & 0.733 && 0.820 & 0.695 && 0.893 & 0.609 & 0.724 && 0.821 & 0.699\\
		 & AML && 0.940 & 0.615 & 0.743 && NA & NA && 0.924 & 0.609 & 0.734 && NA & NA\\
		 & BERTMap && 0.966 & 0.606 & 0.745 && 0.919 & 0.876 && 0.941 & 0.724 & 0.818 && 0.963 & 0.941\\ \midrule
		 \multirow{4}{5em}{\footnotesize SNOMED-NCIT (Neoplas)} & EditSim && 0.815 & 0.709 & 0.759 && 0.900 & 0.876 && 0.775 & 0.713 & 0.743 && 0.900 & 0.876 \\
		 & LogMap && 0.823 & 0.547 & 0.657 && 0.824 & 0.747 && 0.783 & 0.547 & 0.644 && 0.821 & 0.743\\
		 & AML && 0.747 & 0.554 & 0.636 && NA & NA && 0.696 & 0.552 & 0.616 && NA & NA\\
		 & BERTMap && 0.655 & 0.777 & 0.711 && 0.960 & 0.939 && 0.575 & 0.784 & 0.664 && 0.965 & 0.947\\
		 \bottomrule \\
	\end{tabular}
	\end{adjustbox}
	\vspace{-0.5cm}
	\caption{\footnotesize Results of Equivalence Matching.}
	\label{equiv_match:results}
\end{table*}

The equivalence matching results are shown in Table \ref{equiv_match:results}.
The columns of ``$90\%$ Test Mappings'' and ``$70\%$ Test Mappings'' correspond to the unsupervised and semi-supervised data splitting settings, respectively.
From the global matching results, we can see that OMIM-ORDO (Disease) is the most challenging task (with the lowest average F1), while NCIT-DOID (Disease) is the least challenging.
BERTMap attains the highest F1 on OMIM-ORDO (Disease), NCIT-DOID (Disease), SNOMED-NCIT (Pharm), whereas AML is ranked first on SNOMED-NCIT (Body). 
Surprisingly, the naive EditSim method gets the highest F1 score on SNOMED-NCIT (Neoplas), possibly because the ontologies of this task has relatively less hierarchical information to utilize.
%
For the local ranking results, we do not report results of AML because it has no interface for scoring input class pairs.
BERTMap consistently outperforms  EditSim and LogMap,
which is expected because of the advanced BERT-based ML module.

\subsection{Subsumption Matching}



For subsumption matching, we evaluated the following OM systems (methods)\footnote{\label{bertsubsurl}BERTSubs codes: \url{https://gitlab.com/chen00217/bert_subsumption}; Word2Vec (or OWL2Vec*) + RF codes are in the folder \texttt{Inter\_Ontology/baselines/} of the this repository.}:
\begin{enumerate}
    \item \textbf{Word2Vec + Random Forest (RF)}. This method encodes each class by the average of the token vectors of its label defined by \texttt{rdfs:label}. We use a Word2Vec model \cite{mikolov2013efficient} trained by a Wikipedia English article dump accessed in 2018. Given a subsumption, the vectors of its two classes are concatenated and fed to a RF classifier which outputs a mapping score.
    The classifier is trained by the asserted intra-ontology subsumptions in both ontologies for matching in the unsupervised setting. In the semi-supervised setting, these subsumptions are merged with the training mappings for training.
    \item \textbf{OWL2Vec* + RF}. This method is similar to Word2Vec + RF, except that it encodes each class by an ontology embedding model named OWL2Vec*\cite{chen2021owl2vec} which is a Word2Vec model trained on corpora extracted from the ontology with different kinds of semantics concerned. We tested different corpus settings with the best results reported.
    \item \textbf{BERTSubs with Isolated Class (IC)}. BERTSubs with the IC setting \cite{chen2022contextual} has the same architecture as BERTMap, but it fine-tunes the BERT model by the declared subsumptions in the two ontologies for matching. The current results are based on the labels defined by \texttt{rdfs:label}. We will evaluate the other settings that consider surrounding classes in the new OAEI track.
\end{enumerate}

\begin{table*}[!t]
	\centering
    \begin{adjustbox}{width=0.85\textwidth}
	\begin{tabular}{l l c c c c c c c c c c c c c c c}
		\toprule
		 & &  \phantom{a} & \multicolumn{4}{c}{\textbf{90\% Test Mappings}} & \phantom{a} & \multicolumn{4}{c}{\textbf{70\% Test Mappings}} \\
		 \cmidrule{4-7} \cmidrule{9-12} 
		 \textbf{Task} & \textbf{System} & &  \textbf{MRR} & \textbf{H@1} & \textbf{H@5} & \textbf{H@10} && \textbf{MRR} & \textbf{H@1} &
		 \textbf{H@5} & \textbf{H@10} \\ \midrule
		 \multirow{3}{8em}{\footnotesize \text{OMIM-ORDO} (Disease)} & Word2Vec+RF && $0.191$ & $0.106$ & $0.223$ & $0.362$ && $0.193$ & $0.110$ & $0.233$ & $0.315$ \\
		  & OWL2Vec*+RF && $0.270$ & $0.160$ & $0.362$  & $0.521$  && $0.284$ & $0.151$  & $0.411$ & $0.534$ \\
		  & BERTSubs (IC) && $0.299$ & $0.108$ & $0.473$ & $0.613$  && $0.295$& $0.139$ &$0.472$ &$0.667$ \\
		 \midrule
		 \multirow{3}{5em}{\footnotesize \text{NCIT-DOID} (Disease)} & Word2Vec+RF &&$0.306$&$0.206$&$0.390$&$0.510$& &$0.363$ & $0.263$ & $0.448$ & $0.566$\\
		 & OWL2Vec*+RF &&$0.388$&$0.285$&$0.485$&$0.604$&& $0.422$&$0.315$ &$0.524$ &$0.647$ \\
		 & BERTSubs (IC) && $0.601$&$0.460$&$0.777$&$0.877$& &  $0.618$ & $0.496$ &$0.758$ &$0.862$ \\
		 \midrule
		 \multirow{3}{5em}{\footnotesize  \text{SNOMED-FMA} (Body)} & Word2Vec+RF  && $0.558$&$0.415$&$0.731$&$0.850$ && $0.629$ & $0.503$ & $0.792$ & $0.886$ \\
		 & OWL2Vec*+RF && $0.668$ & $0.540$ & $0.836$ & $0.911$ && $0.743$ & $0.626$ & $0.900$ & $0.944$\\
		 & BERTSubs (IC) && $0.589$ & $0.422$ & $0.816$ & $0.939$ &&  $0.622$ & $0.490$ &$0.788$ &$0.878$ \\
		 \midrule
		 \multirow{3}{5em}{\footnotesize \text{SNOMED-NCIT} (Pharm)} & Word2Vec+RF && $0.488$ & $0.335$ & $0.687$ & $0.852$ && $0.526$ & $0.402$ & $0.663$ & $0.834$ \\
		 & OWL2Vec*+RF && $0.524$ & $0.364$ & $0.738$ & $0.870$ &&  $0.579$ & $0.446$ & $0.747$ & $0.893$  \\
		 & BERTSubs (IC) && $0.504$ & $0.321$ & $0.762$ & $0.920$ && $0.476$ & $0.281$ &$0.715$ &$0.900$ \\
		 \midrule
		 \multirow{3}{5em}{\footnotesize \text{SNOMED-NCIT} (Neoplas)} & Word2Vec+RF && $0.512$ & $0.368$ & $0.694$	& $0.834$  && $0.577$ & $0.433$ &$0.773$ &$0.880$  \\
		 & OWL2Vec*+RF && $0.603$ & $0.461$ & $0.782$ & $0.860$ && $0.666$ & $0.547$ & $0.827$ & $0.880$ \\
		 & BERTSubs (IC) && $0.530$ & $0.333$ & $0.786$ & $0.948$ && $0.638$ &$0.463$ &$0.859$ &$0.953$ \\
		 \bottomrule \\
	\end{tabular}
	\end{adjustbox}
	\vspace{-0.5cm}
	\caption{\footnotesize Results of Subsumption Matching.}
	\label{subs_match:results}
\end{table*}


\noindent The setting for negative candidates is the same as in equivalence matching ($N$ is set to $50$; RandomSample is used only when IDFSample or NeighbourSample outputs less than $N$ candidates).
The results are shown in Table \ref{subs_match:results}.
We can find that OWL2Vec* leads to better performance than Word2Vec in all the five tasks when their class embeddings are fed to RF. 
BERTSubs (IC) has higher scores than OWL2Vec* + RF on tasks of OMIM-ORDO and NCIT-DOID for all the four metrics; while on SNOMED-FMA (Body), SNOMED-NCIT (Pharm) and SNOMED-NCIT (Neoplas), BERTSubs (IC) has lower MRR and H@1 scores, but it often has higher H@10 scores than OWL2Vec* + RF.
We can also observe that the results under the semi-supervised setting are usually better than their correspondences under the unsupervised setting, which matches our assumption that adding some training mappings bridges the gap between the intra-ontology subsumptions for training and the inter-ontology subsumptions (mappings) for testing.
Meanwhile, we can find that subsumption matching by BERTSubs (IC) has much lower MRR and H@1 than equivalence matching by BERTMap in each task. Although BERTSubs (IC) only uses one class label, this in some degree verifies that subsumption matching is more challenging. 

\section{Conclusion \& Discussion}




In this paper, we proposed evaluation resources for five biomedical OM tasks that consider both equivalence matching and subsumption matching, with many new features for supporting the evaluation and development of both ML-based and non-ML-based OM systems.
The quality of the reference mappings is ensured by selecting reliable mapping sources (e.g., the human curated mappings from Mondo) and pruning the ontologies.
Subsumption reference mappings are constructed from equivalence reference mappings, where a class deletion algorithm is employed to prevent OM systems from directly inferring the subsumptions through the equivalence mappings. 
We also proposed a comprehensive evaluation framework which includes local ranking and global matching, providing metrics from various perspectives, as well as unsupervised and semi-supervised mapping splitting settings.
Several typical OM systems have been evaluated to demonstrate the application of these resources
and some interesting results and observations have been reported.

While we only constructed datasets for five OM tasks, the resource construction approach
is reproducible for constructing more datasets from Mondo and UMLS for different tasks and settings. 
Most of our techniques, such as category-specific ontology pruning, subsumption mapping construction, and negative candidate generation, are also applicable to general OWL ontologies beyond the biomedical domain, and other tasks beyond OM such as ontology completion.

As for the evaluation, bringing in local ranking amends some key features not properly considered in previous works, thus forming a more comprehensive evaluation framework on assessing both OM systems and mappings.
First, most OM systems, especially those ML-based, rely on a mapping scoring module as well as some other modules for mapping searching (e.g., task blocking, candidate mapping selection and mapping repair).
If an OM system often performs well in local ranking but performs poorly in global matching, then the mapping searching modules need to be debugged and improved. 
Second, even when reference mappings are rather incomplete, local ranking can still provide a fair comparison, especially towards the mapping scoring module, whereas global matching will underestimate Precision of an OM system that has good Recall. 
Actually, local ranking itself simulates some real-world OM applications, such as querying a list of matched classes in a target ontology for a given class in a source ontology.
Third, when many representative OM systems attain high ranking scores but low matching scores on the same set of reference mappings, it is likely that the reference mappings themselves are not complete. 



We are running a new \track track in the OAEI 2022 edition with the proposed datasets. This new track is superseding the current OAEI \textit{largebio} and \textit{phenotype} tracks and, among other objectives, aims at attracting more ML-based systems to the OAEI, which has been highlighted as a key challenge within the OM community.
We will also consider adapting our evaluation framework into MELT (Matching EvaLuation Toolkit) \cite{Hertling2019MELTM}, especially the MELT-ML module for ML-based OM systems, to hold a public evaluation for the OM participants.
Meanwhile, we will also develop and extend our current OM systems BERTMap and BERTSubs based on these new resources, and further consider feeding high-quality system output mappings to the UMLS and Mondo communities. 


\section*{Acknowledgments}
This work was supported by the SIRIUS Centre for Scalable Data Access (Research Council of Norway, project 237889), eBay, Samsung Research UK, Siemens AG, and the EPSRC projects OASIS (EP/S032347/1), UK FIRES (EP/S019111/1) and ConCur (EP/V050869/1). We would like to to thank 
the Mondo team, especially Nicolas Matentzoglu 
and Joe Flake, for their great help in creating the Mondo datasets.




%
%
%
\bibliographystyle{splncs04}
\bibliography{ref}

\end{document}